%% file: main.tex
\documentclass[11pt,letterpaper]{article}
\usepackage[T1]{fontenc}
\usepackage{amsmath,amssymb,amsthm,microtype,xcolor}
\usepackage[margin=1.05in]{geometry}
\usepackage{float,needspace}
\usepackage[noend]{algpseudocode}
\floatstyle{ruled}
\newfloat{algorithm}{tbp}{loa}
\floatname{algorithm}{Algorithm}
\definecolor{paperlink}{RGB}{0,0,114}
\usepackage[colorlinks=true,allcolors=paperlink]{hyperref}
\hypersetup{pdftitle={Beyond blocking in bandits with switching costs},pdfauthor={}}
\newtheorem{theorem}{Theorem}
\newtheorem{lemma}[theorem]{Lemma}
\numberwithin{equation}{section}
\newcommand{\E}{\mathbb E}
\newcommand{\Pp}{\mathbb P}
\newcommand{\ind}{\mathbf1}
\DeclareMathOperator{\Cov}{Cov}
\DeclareMathOperator{\TV}{TV}
\DeclareMathOperator{\KL}{KL}
\DeclareMathOperator{\clip}{clip}

\title{Optimal Bandits with Switching Costs: $(T/\log(T))^{2/3}$ Regret}
\author{Chenyu Gan\thanks{Qiuzhen College, Tsinghua University. \texttt{gancy22@mails.tsinghua.edu.cn}}}\date{}
\begin{document}
\maketitle
\begin{abstract}
\input{abstract}
\end{abstract}
\input{introduction}
\input{related}
\input{algorithm-ideas}
\input{algorithm-block}
\input{algorithm-sample}
\input{algorithm-regret}
\input{bandit-lower}
\section*{Acknowledgement}
The author would like to thank Yuval Peres for helpful discussions on this problem.
\section*{AI disclosure}
Large language models were used to assist with calculations and refinement of the analysis. AI tools were also used for proof review and revision of the exposition. The author verified the correctness and originality of all content, including references.
\bibliographystyle{alpha}
\bibliography{references}
\end{document}

%% file: abstract.tex
We study adversarial bandits with switching costs. For~$K$ arms,~$T$ rounds, unit switching costs, and oblivious losses in~$[0,1]$, we prove that the minimax regret is
\begin{equation*}
 \Theta\!\left(\min\left\{T,
 \frac{K^{1/3}T^{2/3}}{\log^{2/3}(2+T/K)}\right\}\right)\,.
\end{equation*}
This shows that simple blocking, which plays one arm throughout each block, is suboptimal. Our algorithm makes at most a constant number of switches at random times within each block. The key idea for the improvement is to extract more information per switch by giving greater weight to observations near switching times. We also introduce detours, during which the learner temporarily plays a newly sampled arm. Larger observed losses make a detour more likely. These detours cancel temporal correlation terms induced by blocking while preserving the observations used in the estimates. A matching lower bound follows from a refined analysis of a modified multiscale random walk construction.

%% file: introduction.tex
\section{Introduction}\label{sec:introduction}
Switching costs arise when changing a decision consumes time, energy or communication resources. In wireless monitoring, a device observes only the channels to which it is assigned, and changing channels interrupts reception~\cite{le14}. In edge computing, a server can store only a few prediction models; replacing a model requires downloading another from the cloud~\cite{shi25}. Related costs occur in wireless network selection and mobile crowdsensing~\cite{steiger23}. In these applications, exploration requires changing the current action, so acquiring information itself incurs a cost. Bandits with switching costs capture this interaction between partial feedback and the cost of changing actions. We study the adversarial setting, in which the loss sequence is fixed in advance by an oblivious adversary.

Fix integers~$T\ge1$ and~$K\ge2$. An oblivious adversary chooses~$\ell=(\ell_1,\ldots,\ell_T)\in([0,1]^K)^T$. At round~$t$, a learner~$\mathcal A$, knowing~$T,K$, chooses~$A_t\in[K]$ from past observations and private randomness, incurs~$\ell_{t,A_t}$, and observes only this loss. Define its expected regret and number of switches by
\begin{align}
 R_T(\mathcal A,\ell)&=\E_{\mathcal A}\sum_{t=1}^T\ell_{t,A_t}-\min_i\sum_{t=1}^T\ell_{t,i},\label{eq:regret}\\
 S_T(\mathcal A,\ell)&=\E_{\mathcal A}\sum_{t=2}^T\ind_{\{A_t\ne A_{t-1}\}}\,.
 \label{eq:switches}
\end{align}
With unit switching costs and no initial switching charge, the minimax value is
\begin{equation}\label{eq:cost-value}
 \mathcal V_{T,K}=\inf_{\mathcal A}
       \sup_\ell\bigl(R_T(\mathcal A,\ell)+S_T(\mathcal A,\ell)\bigr)\,.
\end{equation}

For~$K\le T$, Audibert and Bubeck~\cite{ab09} provided an algorithm that achieves $O(\sqrt{KT})$ regret for the classical adversarial bandit problem. This result, combined with a simple blocking strategy, gives the best previous upper bound $O(K^{1/3}T^{2/3})$. The best previous lower bound is~$\Omega(\min\{T,K^{1/3}T^{2/3}/\log T\})$, achieved by Dekel, Ding, Koren and Peres~\cite{ddkp}. In this work, we determine the minimax value up to universal constants.

\begin{theorem}\label{thm:cost}
There are universal constants~$c,C>0$ such that, for all~$T\ge1$ and~$K\ge2$,
\begin{equation}\label{eq:main-cost}
 cG_{T,K}\le\mathcal V_{T,K}\le CG_{T,K},\qquad
 G_{T,K}=\min\left\{T,
 \frac{K^{1/3}T^{2/3}}{\log^{2/3}(2+T/K)}\right\}\,.
\end{equation}
\end{theorem}

\paragraph{Simple blocking is suboptimal.}
The most interesting thing about this result is the upper bound improvement. Our result suggests that simple blocking strategy is not optimal. 
This disproves the claim of~\cite{rsc} that Tsallis-Switch is minimax optimal within constants. Their deduction omits the logarithmic denominator in the lower bound of~\cite{ddkp}. 

Our learner makes at most four switches at random times inside each block. It compares two arms using loss averages over dyadic windows around a random switching time in each block. We use an estimator that gives greater weight to observations near a switch. 
The blocking strategy also induces some correlations between losses in the same block. So we further introduce a mechanism called detour, during which the learner plays a newly sampled arm. This mechanism cancels the temporal correlation terms while preserving the observations used in the comparisons. Finally, we use Tsallis regularization to combine these estimates across blocks. Section~\ref{sec:upper} gives the algorithm and its analysis.

The matching lower bound refines the multiscale construction of~\cite{ddkp}. Specifically, we first divide rounds into $K$ blocks and run an independent multiscale random walk in each block. Section~\ref{sec:bandit-lower} gives the proof.

%% file: related.tex
\section{Related work}\label{sec:related}

\paragraph{Bandits with switching costs.}
Agrawal, Hegde and Teneketzis studied switching costs in stochastic allocation and obtained asymptotically efficient policies~\cite{agrawal88}. For adversarial bandits, Arora, Dekel and Tewari gave a general mini-batching reduction against bounded-memory adversaries~\cite{adt12}, yielding~$O((K\log K)^{1/3}T^{2/3})$ regret with switching costs. Cesa-Bianchi, Dekel and Shamir proved an~$\Omega(T^{2/3})$ lower bound when losses are allowed to be unbounded~\cite{cds13}. Dekel, Ding, Koren and Peres established the lower bound for losses in~$[0,1]$, up to logarithmic factors~\cite{ddkp}.

The best previously known upper bound is achieved by Altschuler and Talwar~\cite{at}. They obtained~$O(K^{1/3}T^{2/3})$ adversarial regret by combining the $O(\sqrt{KT})$ regret algorithm in classical adversarial  bandit problem with a simple blocking strategy. Later, Rouyer, Seldin and Cesa-Bianchi introduced Tsallis-Switch, obtaining the same adversarial regret and simultaneous guarantees in the stochastic reward setting~\cite{rsc}. Amir, Azov, Koren and Livni strengthened the latter guarantees~\cite{aakl}. Simchi-Levi and Xu also studied switching constraints for stochastic bandits~\cite{simchilevi23}.

Under full-information feedback, Kalai and Vempala showed that follow-the-perturbed-leader algorithm achieves square-root switching costs~\cite{kv05}. Geulen, V\"ocking and Winkler achieved this rate with the shrinking dartboard algorithm~\cite{gvw}.

Koren, Livni and Mansour generalized switching costs to metric movement costs~\cite{klm17}. Rangi and Franceschetti~\cite{rangi19} and Arora, Marinov and Mohri~\cite{amm19} studied feedback graphs. Cesa-Bianchi et al. characterize which feedback graphs permit square-root dependence on the horizon, up to logarithmic factors~\cite{graphs26}. Cheng, Zhou and Ji quantify the value of additional observations~\cite{cheng23}. Shi, Lin and Jiao consider costly full feedback and simultaneous access to several arms~\cite{shi25}. Dong and Tan study combinatorial bandits with switching costs~\cite{dong24}.

\paragraph{Tsallis-INF regularization.}
Audibert and Bubeck introduced the implicitly normalized forecaster, achieving minimax adversarial regret without the extra~$\sqrt{\log K}$ factor~\cite{ab09}. Abernethy, Lee and Tewari developed its connection with Tsallis entropy~\cite{alt15}. Zimmert and Seldin introduced Tsallis-INF, obtaining optimal adversarial and stochastic guarantees without knowing the regime or horizon~\cite{tsallis}. They also extended their guarantees to utility-based dueling bandits.

Further extensions include the best-of-both-worlds semi-bandit guarantees of Zimmert, Luo and Wei~\cite{zlw19}, decoupled exploration and exploitation by Rouyer and Seldin~\cite{rouyer20}, and linear contextual bandits by Kato and Ito~\cite{kato25}.

%% file: algorithm-ideas.tex
\section{The upper bound}\label{sec:upper}
\subsection{Key ideas}\label{alg:sec-ideas}

In each block, our learner switches between two candidate arms at a random time and may make a detour to another arm. There are two critical mechanisms in our algorithm: (1) use weighted comparisons at dyadic scales (2) make detours to compensate for a ``bad'' arm choice made at the start of the block.

\paragraph{Comparing averages at dyadic scales.}
The multiscale lower bound of Dekel, Ding, Koren and Peres~\cite{ddkp} connects rounds at dyadic distances. This suggests comparing averages on both sides of a switch at distances~$1,2,4,\ldots$, giving equal importance to each scale.

For illustration, consider a block of dyadic length~$B\ge4$, with time indices read modulo~$B$. Choose~$U$ uniformly from~$\{0,\ldots,B-1\}$, play arm~$i$ for the half-block starting at~$U$, and arm~$i'$ for the other half. If~$Y_t$ is the observed loss, the comparison at scale~$m\le B/2$ is
\begin{equation*}
 C_m(U)=\frac1m\sum_{r=0}^{m-1}Y_{U+r}
              -\frac1m\sum_{r=1}^{m}Y_{U-r}\,.
\end{equation*}
Let~$D_i=\sum_{t=0}^{B-1}\ell_{t,i}$ be an arm's block loss. Uniformity of~$U$ gives~$\E_U C_m(U)=(D_i-D_{i'})/B$. Thus, with~$J=\lfloor(\log_2B)/2\rfloor$,
\begin{equation*}
 \widehat D=\frac BJ\sum_{j=0}^{J-1}C_{2^j}(U),
 \qquad \E_U\widehat D=D_i-D_{i'}\,.
\end{equation*}
Observations next to the switch enter every comparison. More distant observations enter the wider windows, each with smaller weight. Equal importance across scales therefore gives greater weight to observations near switches.

To describe what the weights control, write the two losses as
\begin{equation*}
 \ell_{t,i}=c_t+\frac{d_t}{2},\qquad
 \ell_{t,i'}=c_t-\frac{d_t}{2},\qquad
 c_t=\frac{\ell_{t,i}+\ell_{t,i'}}2,\quad
 d_t=\ell_{t,i}-\ell_{t,i'}\,.
\end{equation*}
Here~$c_t$ is the same additive component in both losses, while~$d_t$ is their difference. The dyadic weights bound the estimator's variance of the contribution from~$c_t$ by~$O(B^2/J^2)$. 

\paragraph{Blocking and detours.}
Blocking keeps the same sampled arm over several rounds. The observations at those rounds depend on that shared arm choice, so squaring a block estimate produces temporal cross terms. For a pair of arms, the contribution from $d_t$ to the estimator's variance involves~$d_s d_t$. When averaged over the candidate pairs, these products become covariances: if~$p$ is the distribution supplied by the learning rule at the start of the block, then
\begin{equation*}
 \sum_{i<j}p_ip_j(\ell_{s,i}-\ell_{s,j})(\ell_{t,i}-\ell_{t,j})
 =\Cov_{i\sim p}(\ell_{s,i},\ell_{t,i})\,.
\end{equation*}
These terms are positive when the same arms tend to have higher losses at both rounds. The weighted comparison bound retains a sum of these covariances. The detours supply the opposite sum through a change in the loss actually incurred.

\Needspace{8\baselineskip}
We use the observed loss to decide when to leave the current arm temporarily. Larger losses make a detour more likely. The basic calculation starts with~$I\sim p$ and replaces it by an independent draw from~$p$ with probability~$\ell_{s,I}$. If~$I'$ is the resulting arm, then
\begin{align*}
 \E[\ell_{t,I'}-\ell_{t,I}]
 &=\sum_i p_i\ell_{s,i}
       \left(\sum_jp_j\ell_{t,j}-\ell_{t,i}\right)\notag\\
 &=-\Cov_{i\sim p}(\ell_{s,i},\ell_{t,i})\,.
\end{align*}
Thus, when a large loss at~$s$ predicts a large loss at~$t$, resampling more often after large losses reduces the expected loss at~$t$. This produces exactly the negative covariance correction needed in the regret analysis.

%% file: algorithm-block.tex
\subsection{The algorithm}\label{alg:sec-many}
The learner divides time into blocks. At the beginning of a block it chooses two candidate arms and fixes a base path using those arms. Inside the block it may temporarily replace the base arm after observing its loss at one uniformly chosen round, with larger observed losses giving a larger triggering probability. The observations used to estimate the difference between the candidates are protected from these replacements. The difference estimator therefore retains its prescribed distribution, while the temporary replacements contribute a signed correction to actual loss.

\paragraph{Block routine.}
Fix a power of two~$4\le B\le T$ and divide the horizon into $Q=\lceil T/B\rceil$ consecutive blocks of length~$B$. Run the block routine separately in each block and truncate the resulting action sequence at~$T$.
Within a block, use local rounds~$t=0,\ldots,B-1$, and write~$\ell_t$ for the loss vector and~$A_t$ for the action.

At the start of the block, sample arms~$i\le j$ and, if~$i<j$, a split length~$k\in\{1,\ldots,B-1\}$. Their sampling rule is specified later. Independently draw~$U$ uniformly from $\{0,\ldots,B-1\}$. For~$i<j$, define the base arm by
\begin{equation*}
 a_t=
 \begin{cases}
 i,&(t-U)\bmod B<k,\\
 j,&(t-U)\bmod B\ge k.
 \end{cases}
\end{equation*}
For~$i=j$, set~$a_t=i$ throughout the block. The sequence~$(a_t)$ specifies which arm to play when there is
no detour. 

Set~$M=\log_2 B$,~$J=\lfloor M/2\rfloor$ and~$m_j=2^j$ for~$0\le j<J$. Define the weights by
\begin{equation}\label{alg:kappa}
 \kappa_r=\sum_{j=0}^{J-1}
 \frac{\ind_{\{r<m_j\}}-\ind_{\{r\ge B-m_j\}}}{m_j}
 \quad(0\le r<B),
 \qquad
 w_t=\kappa_{(t-U)\bmod B}\,.
\end{equation}
Write~$m_*=m_{J-1}\le\sqrt B/2$. The nonzero entries of~$\kappa$ are its first and last~$m_*$ entries, and its positive and negative masses are both~$J$. These weights are used to construct the loss estimator.

As discussed before, we introduce a detour to cancel out the second order correlation terms due to the blocking. 
At the start of the block, draw~$\tau$ uniformly from~$\{0,\ldots,B-1\}$. If~$w_\tau=0$, then after observing~$Y_\tau=\ell_{\tau,a_\tau}$, trigger a detour with probability~$Y_\tau$. When a detour is triggered, independently draw a destination arm~$C$ from the distribution~$p$ fixed at the start of the block and a requested duration~$R\in\mathbb Z_{\ge0}$ from
\begin{equation}\label{alg:duration}
 \Pp(R=r)=
 \begin{cases}
 \displaystyle
 1-\frac14\sum_{j=0}^{J-1}
       \frac{(j+1)(m_j-1)}{m_j^2},
       &r=0,\\[6pt]
 \displaystyle
 \frac14\sum_{j=0}^{J-1}\frac{j+1}{m_j^2}
       \ind_{\{r<m_j\}},
       &r\ge1.
 \end{cases}
\end{equation}
This distribution is supported on~$\{0,\ldots,m_*-1\}$.

If a detour is triggered, define
\begin{equation}\label{alg:safe-endpoint}
 h=\max\left\{u\in\{\tau,\ldots,B-1\}:
 w_s=0,\ a_s=a_\tau
 \text{ for every }s=\tau,\ldots,u\right\}.
\end{equation}
This is the latest round that the detour may reach. It protects the estimator by requiring the detour to end before an observation with nonzero weight, and also before the next base-arm change.
More specifically, a detour accepted at~$\tau$ plays arm~$C$ on rounds $\tau+1,\ldots,\min\{h,\tau+R\}$ and then resumes the base arm.
Thus every observation with~$w_t\ne0$ comes from~$a_t$.

\begin{algorithm}[H]
\caption{A block with protected observations}\label{alg:block}
\begin{algorithmic}[1]
\Require Block length~$B$, candidate arms~$i,j$, split length~$k$,
destination distribution~$p$
\State Draw~$U$ and~$\tau$ independently and uniformly from~$\{0,\ldots,B-1\}$
\State For each~$t$, set~$a_t\gets i$ if~$(t-U)\bmod B<k$, and~$a_t\gets j$ otherwise
\State Compute~$\kappa$ from~\eqref{alg:kappa} and set~$w_t\gets\kappa_{(t-U)\bmod B}$ for every~$t$
\State $W\gets0$, $e\gets-1$, $C\gets1$
\For{$t=0,\ldots,B-1$}
 \If{$t\le e$}
  \State $A_t\gets C$
 \Else
  \State $A_t\gets a_t$
 \EndIf
 \State Play~$A_t$ and observe~$Y_t=\ell_{t,A_t}$
 \State $W\gets W+w_tY_t$
 \If{$t=\tau$ and $w_t=0$}
  \If{a Bernoulli draw of mean~$Y_t$ equals one}
   \State Independently draw~$C\sim p$ and~$R$
          from~\eqref{alg:duration}
   \State Compute~$h$ from~\eqref{alg:safe-endpoint}
   \State $e\gets\min\{h,\tau+R\}$
  \EndIf
 \EndIf
\EndFor
\State \Return $W$
\end{algorithmic}
\end{algorithm}

%% file: algorithm-sample.tex
\paragraph{Sampling and update.}
Let~$\Delta_K=\{x\in[0,1]^K:\sum_i x_i=1\}$ be the probability simplex and define the maximizer of the Tsallis potential
\begin{equation}\label{sc:tsallis-potential}
 p(y)=\arg\max_{x\in\Delta_K}\left\{\langle y,x\rangle+2\sum_i\sqrt{x_i}\right\}\,.
\end{equation}
For each block $b$, set~$p_b=p(y_b)$ and~$\nu_b=(p_b+\mathbf1/K)/2$, where~$\mathbf1$ denotes the all-one vector. The update rule for $y_b$ is specified later. Draw base arm pairs independently from~$p_b$ and~$\nu_b$. A distinct unordered pair~$\{i,j\}$,~$i<j$, has probability and base proportion
\begin{equation}\label{alg:pair-law}
 \omega_{b,ij}=p_{b,i}\nu_{b,j}+p_{b,j}\nu_{b,i},
 \qquad \alpha_{b,ij}=\frac{p_{b,i}\nu_{b,j}}{\omega_{b,ij}}\,.
\end{equation}
For a distinct pair, set $k=k_{b,ij}=\max\{1, \lfloor B\alpha_{b,ij}\rfloor\}$.

For~$1\le k<B$, the normalizing mass is
\begin{equation}\label{alg:S}
 S_k=\sum_{r=0}^{k-1}\kappa_r
     \,.
\end{equation}
For a distinct pair, let~$W$ be returned by Algorithm~\ref{alg:block} with parameters $(B,i,j,k,p_b)$. Set~$\widehat D_{ij}=BW/S_k$ and
\begin{equation*}
 g_{b,a}=\begin{cases}
 p_{b,j}\widehat D_{ij}/\omega_{b,ij},&a=i,\\
 -p_{b,i}\widehat D_{ij}/\omega_{b,ij},&a=j,\\
 0,&a\notin\{i,j\}.
 \end{cases}
\end{equation*}
For a singleton, set~$g_b=\mathbf0$.

\begin{algorithm}[H]
\caption{The $K$-arm learner}\label{alg:many}
\begin{algorithmic}[1]
\Require Horizon~$T$, number of arms~$K\ge2$, dyadic block length~$4\le B\le T$
\State Set $y_1\gets\mathbf 0$, $\eta\gets\log^2B/(432\sqrt K B^2)$. Pad the final block with zeros
\For{each block $b=1,\ldots, Q$}
 \State $p_b\gets p(y_b)$ from~\eqref{sc:tsallis-potential}, $\nu_b\gets(p_b+\mathbf1/K)/2$
 \State Independently draw candidates from~$p_b$ and~$\nu_b$, label them~$i\le j$
 
 \If{$i=j$}
  \State Set~$k=B$
 \Else
  \State Compute~$\omega_{b,ij},\alpha_{b,ij}$ from~\eqref{alg:pair-law}
  \State $k\gets\max\{1,\lfloor B\alpha_{b,ij}\rfloor\}$
 \EndIf
 \State $W\gets$ Algorithm~\ref{alg:block} with inputs $(B,i,j,k,p_b)$
 \State $g_b\gets\mathbf0$
 \If{$i\ne j$}
  \State $\widehat D_{ij}\gets BW/S_k$
  \State $g_{b,i}\gets p_{b,j}\widehat D_{ij}/\omega_{b,ij}$; $g_{b,j}\gets-p_{b,i}\widehat D_{ij}/\omega_{b,ij}$
 \EndIf
 \State $y_{b+1}\gets y_b-\eta g_b$
\EndFor
\State Truncate the action path at~$T$
\end{algorithmic}
\end{algorithm}

%% file: algorithm-regret.tex
\subsection{Estimates within a block}\label{alg:sec-estimates}

Fix the loss table and pad the last block with zeros. Write~$\ell_{b,t,i}$ for the loss of arm~$i$ at local round~$t$ in block~$b$ and~$A_{b,t}$ for the arm played there. Set~$D_{b,i}=\sum_{t=0}^{B-1}\ell_{b,t,i}$ and~$D_b=(D_{b,i})_{i=1}^K$. Let~$\mathcal F_{b-1}$ be the learner's history before block~$b$ and write~$\E_b$ for expectation conditional on that history. In the calculations for a fixed block, we suppress~$b$ on the losses, actions, block totals, probabilities, rounded lengths, and update. In particular,~$p,\nu$ and the quantities~$\omega_{ij},\alpha_{ij},k_{ij}$ are fixed under this conditioning. All sums over rounds within a block use~$0\le t<B$.

\paragraph{Estimating the loss difference.}
For distinct candidate arms~$i<j$ and a split~$1\le k<B$, the weight sum and the estimate are~$W=\sum_t w_tY_t$ and~$\widehat D_{ij}=BW/S_k$, with~$S_k$ from~\eqref{alg:S}.

\begin{lemma}\label{alg:comparison}
The normalizing mass is
\begin{equation}\label{alg:normalizing-mass}
 S_k=\sum_{j'=0}^{J-1}\frac{\min\{k,B-k,m_{j'}\}}{m_{j'}}\,.
\end{equation}
Conditional on the pair~$\{i,j\}$, Algorithm~\ref{alg:block} gives
\begin{equation}\label{alg:mean-range}
 \E_b[\widehat D_{ij}\mid\{i,j\}]=D_i-D_j,
 \qquad |\widehat D_{ij}|\le BJ\,.
\end{equation}
\end{lemma}

\begin{proof}
At scale~$m$, the indices~$0,\ldots,k-1$ contain~$\min\{k,m\}$ positive weights and~$(k-B+m)_+$ negative weights, where~$x_+=\max\{x,0\}$. Since~$m\le B/2$, their difference is~$k$ for~$k\le m$,~$m$ for~$m\le k\le B-m$, and~$B-k$ for~$k\ge B-m$. Dividing by~$m$ and summing over scales gives~\eqref{alg:normalizing-mass}.

At each~$t$ with~$w_t\ne0$, the detour rule gives~$Y_t=\ell_{t,a_t}$. Conditional on the pair,~$(t-U)\bmod B$ is uniform. Thus
\begin{align*}
 \E_b[w_tY_t\mid\{i,j\}]
 &=\frac1B\left(\ell_{t,i}\sum_{r<k}\kappa_r
                 +\ell_{t,j}\sum_{r\ge k}\kappa_r\right)
 =\frac{S_k}{B}(\ell_{t,i}-\ell_{t,j})\,.
\end{align*}
The last equality uses~$\sum_r\kappa_r=0$. Summing and multiplying by~$B/S_k$ gives the mean. The scale~$m_0=1$ contributes one to~\eqref{alg:normalizing-mass}, so~$S_k\ge1$. Since the positive and negative weights each have mass~$J$ and~$0\le Y_t\le1$, we have~$|\widehat D_{ij}|=B|W|/S_k\le BJ$.
\end{proof}

\paragraph{Second moment of the loss estimate.}
For a vector~$z=(z_r)_{r=0}^{B-1}$, write~$\operatorname{circ}(z)$ for the circulant matrix~$(z_{(t-u)\bmod B})_{t,u}$. The Fourier transform at frequency~$n$ is
\begin{equation*}
 \widehat z(n)=\sum_{r=0}^{B-1}z_r e^{-2\pi\mathrm i nr/B}
 \qquad(0\le n<B)\,.
\end{equation*}
Then the vector~$(e^{2\pi\mathrm i nu/B})_u$ is an eigenvector of~$\operatorname{circ}(z)$ with eigenvalue~$\widehat z(n)$. These vectors form an orthogonal basis. Consequently~$\|\operatorname{circ}(z)\|=\max_n|\widehat z(n)|$, where the norm is the Euclidean operator norm.

\begin{lemma}[Convolution norm]\label{alg:filter-norm}
We have~$\|\operatorname{circ}(\kappa)\|\le6$.
\end{lemma}

\begin{proof}
Let~$\kappa^{(j)}=\big((\mathbf1_{\{r<m_j\}}-\mathbf1_{\{r\ge B-m_j\}})/m_j\big)_{0\le r<B}$ be the vector formed by the $j$-th summand in~\eqref{alg:kappa}. For~$n\ne0$, set~$x=\pi n/B$. A direct computation gives
\begin{align*}
 |\widehat{\kappa^{(j)}}(n)|
 &=\frac{|1-e^{2\mathrm i m_jx}|\,|1-e^{-2\mathrm i m_jx}|}
         {m_j|1-e^{-2\mathrm i x}|}
 =\frac{2\sin^2(m_jx)}{m_j\sin x}\\
 &\le 2\min\{m_j\sin x,(m_j\sin x)^{-1}\}\,.
\end{align*}
Here~$|\sin(m_jx)|\le\min\{m_j\sin x,1\}$ follows by applying~$|\sin(2z)|\le2|\sin z|$ repeatedly. 

Choose~$l\ge0$ with~$a=2^l\sin x\in[1/2,1]$. Then
\begin{align*}
 |\widehat\kappa(n)|&\le\sum_{j=0}^{J-1}|\widehat{\kappa^{(j)}}(n)|\\
 &\le 2\sum_{j=0}^{l}2^j\sin x
       +2\sum_{j=l+1}^\infty (2^j\sin x)^{-1}\\
 &=2(2^{l+1}-1)\sin x+\frac{2^{1-l}}{\sin x}
 \le 2(2a+a^{-1})\le6\,.
\end{align*}
The last inequality follows from~$3-2a-a^{-1}=(1-a)(2a-1)/a\ge0$. At zero frequency,~$\widehat\kappa(0)=0$ because each scale has signed sum zero. Hence~$\|\operatorname{circ}(\kappa)\|\le6$.
\end{proof}

For a split~$1\le k<B$, define~$f_r=\kappa_r(2\ind_{\{r<k\}}-1)$.
We also define
\begin{align}\label{alg:H}
 H_r^+&=\sum_{j=0}^{J-1}\frac{j+1}{m_j}\left(1-\frac r{m_j}\right)_+
 \qquad(r\ge0),\notag\\
 H_r&=H_r^++H_{B-r}^+\qquad(0\le r<B)\,.
\end{align}

\begin{lemma}[Gap autocorrelation]\label{alg:gap-kernel}
We have~$H_0<4$ and, for every~$z\in\mathbb R^B$,
\begin{equation*}
 \|\operatorname{circ}(f)^\top z\|^2
 \le25z^\top\operatorname{circ}(H)z\,.
\end{equation*}
\end{lemma}

\begin{proof}
Let~$f_r^{(j)}=\kappa_r^{(j)}(2\ind_{\{r<k\}}-1)$. Then~$f=\sum_{j<J}f^{(j)}$. Notice that, for~$0\le r<B$,
\begin{equation*}
 m_j f_r^{(j)}=
 \begin{cases}
 \ind_{\{0\le r<k\}}-\ind_{\{k\le r<m_j\}}
                   +\ind_{\{B-m_j\le r<B\}},&k<m_j,\\
 \ind_{\{0\le r<m_j\}}+\ind_{\{B-m_j\le r<B\}},
                                      &m_j\le k\le B-m_j,\\
 \ind_{\{0\le r<m_j\}}-\ind_{\{B-m_j\le r<k\}}
                   +\ind_{\{k\le r<B\}},&k>B-m_j.
 \end{cases}
\end{equation*}
Thus~$|f_r^{(j)}|=1/m_j$ on~$2m_j$ positions and vanishes elsewhere, so~$\sum_r|f_r^{(j)}|=2$. Reading~$r+1$ modulo~$B$, the cyclic differences are
\begin{equation*}
 m_j(f_r^{(j)}-f_{r+1}^{(j)})=
 \begin{cases}
 2\ind_{\{r=k-1\}}-\ind_{\{r=m_j-1\}}-\ind_{\{r=B-m_j-1\}},
                                      &k<m_j,\\
 \ind_{\{r=m_j-1\}}-\ind_{\{r=B-m_j-1\}},
                                      &m_j\le k\le B-m_j,\\
 \ind_{\{r=m_j-1\}}+\ind_{\{r=B-m_j-1\}}-2\ind_{\{r=k-1\}},
                                      &k>B-m_j.
 \end{cases}
\end{equation*}
Since~$2m_j<B$, the jump locations in each case are distinct. Consequently,
\begin{equation*}
 \sum_{r=0}^{B-1}|f_r^{(j)}-f_{r+1}^{(j)}|
 =\begin{cases}
 2/m_j,&m_j\le k\le B-m_j,\\
 4/m_j,&k<m_j\ \text{or}\ k>B-m_j
 \end{cases}
 \le\frac4{m_j}\,.
\end{equation*}
For~$1\le n<B$, again set~$x=\pi n/B$. Then
\begin{align*}
 (1-e^{2\pi\mathrm i n/B})\widehat{f^{(j)}}(n)
 &=\sum_{r=0}^{B-1}(f_r^{(j)}-f_{r+1}^{(j)})
                         e^{-2\pi\mathrm i nr/B},\\
 |\widehat{f^{(j)}}(n)|
 &\le\min\left\{\sum_r|f_r^{(j)}|,
       \frac{\sum_r|f_r^{(j)}-f_{r+1}^{(j)}|}
            {2\sin x}\right\}
 \le\min\left\{2,\frac2{m_j\sin x}\right\}\,.
\end{align*}
At scale zero,~$f^{(0)}$ has value one at~$0$ and~$B-1$, so~$|\widehat{f^{(0)}}(n)|=2|\cos x|$.

Notice that, the overlap of the cyclic intervals~$\{0,\ldots,m_j-1\}$ and its translate by~$r$ has size~$(m_j-r)_++(m_j-(B-r))_+$, hence
\begin{align*}
 H_r&=\sum_{j<J}\frac{j+1}{m_j^2}
       \sum_{u=0}^{B-1}\ind_{\{u<m_j\}}
                    \ind_{\{(u+r)\bmod B<m_j\}},\\
 \widehat H(n)
 &=\sum_{j<J}\frac{j+1}{m_j^2}
       \left(\sum_{u=0}^{m_j-1}e^{2\pi\mathrm i nu/B}\right)
       \left(\sum_{t=0}^{m_j-1}e^{-2\pi\mathrm i nt/B}\right)\\
 &=\sum_{j<J}(j+1)
       \left|\frac1{m_j}\sum_{t=0}^{m_j-1}e^{-2\pi\mathrm i nt/B}\right|^2\,\\
       &=
 \begin{cases}
 J(J+1)/2,&n=0,\\[2pt]
 \displaystyle\sum_{j<J}(j+1)
       \left(\frac{\sin(m_jx)}{m_j\sin x}\right)^2,&n\ne0.
 \end{cases}
\end{align*}
Also~$H_0=\sum_{j<J}(j+1)2^{-j}<\sum_{j\ge0}(j+1)2^{-j}=4$.

Next, we show that $|\widehat f(n)|^2\le25\widehat H(n)$ for every $n$. The same Fourier basis diagonalizes both~$\operatorname{circ}(f)\operatorname{circ}(f)^\top$ and~$\operatorname{circ}(H)$, with respective eigenvalues~$|\widehat f(n)|^2$ and~$\widehat H(n)$. This bound therefore implies the claimed quadratic-form inequality.

Since $f$ and $H$ are real valued, and $H_r=H_{B-r}$, we have $|\widehat f(B-n)|=|\widehat f(n)|$ and $\widehat H(B-n)=\widehat H(n)$. So it suffices to prove the inequality for $0\le n\le B/2$.

For $n=0$,~$|\widehat f(0)|\le2J$, and therefore
\begin{equation*}
 \frac{|\widehat f(0)|^2}{\widehat H(0)}
 \le\frac{4J^2}{J(J+1)/2}<8\,.
\end{equation*}

For~$n\ne0$, set $q=\min\left\{J,\left\lfloor\log_2(B/n)\right\rfloor\right\}$.
If~$q=1$ and~$J=1$, then~$|\widehat f(n)|\le2$ and~$\widehat H(n)=1$. If~$q=1<J$, then~$\pi/4<x\le\pi/2$, so we have
\begin{equation*}
 |\widehat f(n)|
 \le2\cos x+\frac2{\sin x}\sum_{j=1}^{J-1}2^{-j}
 \le3\sqrt2,
 \qquad \widehat H(n)\ge1\,.
\end{equation*}
Thus the ratio is at most~$18$ when~$q=1$.

Suppose~$q\ge2$. The tail over~$j\ge q$ is empty when~$q=J$. Otherwise~$x>\pi/2^{q+1}$, and concavity of sine on~$[0,\pi/8]$, together with~$\sin(\pi/8)>3/8$, gives~$\sin x>\sin(\pi/2^{q+1})>3/2^{q+1}$. Hence in both cases
\begin{equation*}
 |\widehat f(n)|
 \le2q+\frac2{\sin x}\sum_{j=q}^{J-1}2^{-j}
 \le2q+\frac83\,.
\end{equation*}
Also~$x\le\pi/2^q$. For~$1\le j\le q-2$, concavity of sine gives~$\sin(m_jx)/(m_j\sin x)\ge2\sqrt2/\pi$, while for~$j=q-1$ it gives a lower bound of~$2/\pi$. Keeping these terms yields
\begin{equation*}
 \widehat H(n)\ge
 1+\frac8{\pi^2}\sum_{j=1}^{q-2}(j+1)+\frac{4q}{\pi^2}
 =1+\frac{4q^2-8}{\pi^2}>\frac{2q^2+1}{5}\,.
\end{equation*}
The last inequality uses~$\pi^2<10$. Finally, since $5(2q^2+1)-(2q+8/3)^2>0$, we have~$|\widehat f(n)|^2<25\widehat H(n)$.
\end{proof}

\begin{lemma}[Second moment of the loss estimate]\label{alg:second-moment}
For a fixed distinct pair~$i<j$ and any split~$1\le k<B$ in Algorithm~\ref{alg:block}, write~$d_t=\ell_{t,i}-\ell_{t,j}$. Then
\begin{equation}\label{alg:moment}
 \E_b[\widehat D_{ij}^{\,2}\mid\{i,j\}]
 \le\frac{36B^2}{S_k^2}
   +\frac{50B}{3S_k^2}\sum_{0\le s<t<B}H_{t-s}d_s d_t\,.
\end{equation}
\end{lemma}

\begin{proof}
Write~$c_t=(\ell_{t,i}+\ell_{t,j})/2$ and regard~$c,d$ as vectors indexed by the local rounds. Protection of the weighted observations gives
\begin{equation*}
 W=\sum_t\kappa_{(t-U)\bmod B}c_t
      +\frac12\sum_t f_{(t-U)\bmod B}d_t\,.
\end{equation*}
Use~$(x+y/2)^2\le4x^2+y^2/3$, which follows from~$4x^2+y^2/3-(x+y/2)^2=3(x-y/6)^2$. Average over~$U$ and use~$\operatorname{circ}(\kappa)^\top\mathbf1=0$ to center~$c$. Since both losses lie in~$[0,1]$, we have~$|c_t-1/2|\le(1-|d_t|)/2$. Applying Lemmas~\ref{alg:filter-norm} and~\ref{alg:gap-kernel} gives
\begin{align*}
 \frac{S_k^2}{B}\E_b[\widehat D_{ij}^{\,2}\mid\{i,j\}]
 &\le 4\|\operatorname{circ}(\kappa)^\top c\|^2+\frac13\|\operatorname{circ}(f)^\top d\|^2\\
 &\le144\sum_t(c_t-1/2)^2+\frac{25}{3}d^\top\operatorname{circ}(H)d\\
 &\le \sum_t\left[36(1-|d_t|)^2+\frac{100}3d_t^2\right]
                  +\frac{50}3\sum_{s<t}H_{t-s}d_s d_t\,.
\end{align*}
For the last inequality, the diagonal of the quadratic form contributes~$H_0\sum_t d_t^2$, with~$H_0<4$. The two orders of a distinct pair of times have equal weights because~$H_r=H_{B-r}$ for~$1\le r<B$. Each summand of the first sum is at most~$36$, since~$36(1-x)^2+(100/3)x^2\le36[(1-x)^2+x^2]\le36$ for~$0\le x\le1$. This proves~\eqref{alg:moment}.
\end{proof}

\subsection{Loss of a detour}\label{alg:sec-detours}
We now average over the sampled arms as well as the shift and detour. The comparison loss for the block is~$\langle p,D\rangle=\sum_t\sum_i p_i\ell_{t,i}$.

\begin{lemma}[Detour loss]\label{alg:clock}
When the detour destination has law~$p$, Algorithms~\ref{alg:block} and~\ref{alg:many} satisfy
\begin{equation}\label{alg:clock-loss}
 \E_b\sum_t\ell_{t,A_t}-\langle p,D\rangle
 \le \frac54-\frac1{4B}\sum_{s<t}H_{t-s}
                         \Cov_{i\sim p}(\ell_{s,i},\ell_{t,i})\,.
\end{equation}
\end{lemma}

\begin{proof}
\emph{Base marginal.}
Let~$\widetilde p_i=\Pp(a_t=i\mid\mathcal F_{b-1})$. This is independent of~$t$ by uniformity of~$U$. Before rounding, a pair~$i<j$ contributes mass~$\omega_{ij}\alpha_{ij}=p_i\nu_j$ to arm~$i$ and~$\omega_{ij}(1-\alpha_{ij})=p_j\nu_i$ to arm~$j$. Adding the singleton mass~$p_i\nu_i$ and the contributions from distinct partners gives total mass~$p_i\sum_j\nu_j=p_i$ to arm $i$. After rounding, each pair shifts mass~$\omega_{ij}|k_{ij}/B-\alpha_{ij}|$ between its arms. The rule~$k_{ij}=\max\{1,\lfloor B\alpha_{ij}\rfloor\}$ gives~$1\le k_{ij}<B$ and~$|k_{ij}/B-\alpha_{ij}|\le1/B$. Hence
\begin{equation*}
 \TV(\widetilde p,p)
 =\tfrac12\sum_i|\widetilde p_i-p_i|
 \le\sum_{i<j}\omega_{ij}|k_{ij}/B-\alpha_{ij}|
 \le\frac1B\,.
\end{equation*}
As losses lie in~$[0,1]$, this implies~$\E_b\sum_t\ell_{t,a_t}\le\langle p,D\rangle+B\TV(\widetilde p,p)\le\langle p,D\rangle+1$.

\emph{Loss change before clipping.}
For an integer~$r\ge1$, the distribution in~\eqref{alg:duration} gives
\begin{align}\label{alg:tail}
 \Pp(R\ge r)&=\sum_{a=r}^{\infty}\frac14\sum_{j<J}\frac{j+1}{m_j^2}\ind_{\{a<m_j\}}=\frac14H_r^+,\notag\\
 \mu:=\E R&=\frac14\sum_{j<J}\frac{j+1}{m_j^2}\sum_{r=1}^{m_j-1}(m_j-r)
 \le\frac18\sum_{j<J}(j+1)=\frac{J(J+1)}{16}\,.
\end{align}
Draw~$C\sim p$,~$R$ from~\eqref{alg:duration}, and~$V\sim\operatorname{Unif}[0,1]$ at the start of the block, independently of one another and of the candidate arms,~$U$, and~$\tau$. Use~$V\le\ell_{\tau,a_\tau}$ to decide whether to trigger the detour when~$w_\tau=0$. These draws give the same action law as Algorithm~\ref{alg:block}, since there is no detour before~$\tau$ and hence~$Y_\tau=\ell_{\tau,a_\tau}$.

Define the loss change before clipping by
\begin{equation*}
 Z=\sum_{0\le s<t<B}
   \ind_{\{\tau=s\}}\ind_{\{V\le\ell_{s,a_s}\}}\ind_{\{R\ge t-s\}}
       (\ell_{t,C}-\ell_{t,a_s})\,.
\end{equation*}
Conditional on the base path, independence and the sampling rules give
\begin{align*}
 \E_b[Z\mid(a_u)_{u=0}^{B-1}]
 &=\frac1B\sum_{s<t}\Pp(R\ge t-s)\ell_{s,a_s}
                     (\langle p,\ell_t\rangle-\ell_{t,a_s}),\\
 \E_b Z
 &=\frac1B\sum_{s<t}\Pp(R\ge t-s)
       \sum_i\widetilde p_i\ell_{s,i}
                     (\langle p,\ell_t\rangle-\ell_{t,i})\,.
\end{align*}
For each~$s<t$,
\begin{align*}
 \sum_i p_i\ell_{s,i}(\langle p,\ell_t\rangle-\ell_{t,i})
 &=\langle p,\ell_s\rangle\langle p,\ell_t\rangle
                         -\sum_i p_i\ell_{s,i}\ell_{t,i}\\
 &=-\Cov_{i\sim p}(\ell_{s,i},\ell_{t,i})\,.
\end{align*}
Since~$0\le\ell_{s,i},\ell_{t,i}\le1$,
\begin{equation*}
 \langle p,\ell_t\rangle-1
 \le\ell_{s,i}(\langle p,\ell_t\rangle-\ell_{t,i})
 \le\langle p,\ell_t\rangle\,.
\end{equation*}
The values in the middle therefore lie in an interval of length one. The total variation bound from the first part gives
\begin{equation*}
 \left|\sum_i(\widetilde p_i-p_i)\ell_{s,i}
                  (\langle p,\ell_t\rangle-\ell_{t,i})\right|
 \le\TV(\widetilde p,p)\le\frac1B\,.
\end{equation*}
Using~$\Pp(R\ge t-s)=H_{t-s}^+/4$, we obtain
\begin{align*}
 \left|\E_b Z+\frac1{4B}\sum_{s<t}H_{t-s}^+
                          \Cov_{i\sim p}(\ell_{s,i},\ell_{t,i})\right|
 &\le\frac1{B^2}\sum_{s<t}\Pp(R\ge t-s)\\
 &=\frac1{B^2}\sum_{r=1}^{B-1}(B-r)\Pp(R\ge r)
 \le\frac{\mu}{B}\,.
\end{align*}

\emph{Clipping.}
For~$0\le s<t<B$, define the event
\begin{equation*}
 \mathcal E_{s,t}
 =\{w_u=0,\ a_u=a_s\text{ for every }u=s,\ldots,t\}\,.
\end{equation*}
By~\eqref{alg:safe-endpoint}, a detour triggered after~$s$ reaches~$t$ exactly when~$R\ge t-s$ and~$\mathcal E_{s,t}$ holds. On this event,~$a_t=a_s$. Consequently the actual loss change is
\begin{equation*}
 \sum_{t=0}^{B-1}(\ell_{t,A_t}-\ell_{t,a_t})
 =\sum_{0\le s<t<B}
   \ind_{\{\tau=s\}}\ind_{\{V\le\ell_{s,a_s}\}}\ind_{\{R\ge t-s\}}
   \ind_{\mathcal E_{s,t}}(\ell_{t,C}-\ell_{t,a_s})\,.
\end{equation*}

Fix the candidate arms and their split~$k$, and set~$r=t-s$. Only~$r<m_*$ contributes, since~$R\le m_*-1$. The protected positions are
\begin{equation*}
 \{u\in\{0,\ldots,B-1\}:w_u\ne0\}
 =\{(U+d)\bmod B:-m_*\le d<m_*\}\,.
\end{equation*}
If a protected position lies in~$\{s,\ldots,t\}$, then~$U$ is congruent modulo~$B$ to an integer in~$\{s-m_*+1,\ldots,t+m_*\}$. This interval has~$2m_*+r$ integers, so
\begin{equation*}
 \#\{U:\exists u\in\{s,\ldots,t\},\ w_u\ne0\}\le2m_*+r\,.
\end{equation*}
For distinct candidate arms, the base arm can change only at~$U$ or~$(U+k)\bmod B$. A change at~$U$ inside~$\{s+1,\ldots,t\}$ is already counted, since~$w_U\ne0$. A change at~$(U+k)\bmod B$ inside that interval requires
\begin{equation*}
 U\equiv s+1-k,\ldots,t-k\pmod B,
\end{equation*}
which adds at most~$r$ shifts. For a singleton, the base arm is constant. Thus in both cases
\begin{equation*}
 \#\{U:\mathcal E_{s,t}\text{ fails}\}
 \le2m_*+2r\le4m_*,
 \qquad
 \E_b\ind_{\mathcal E_{s,t}^c}\le\frac{4m_*}{B}\,.
\end{equation*}
The last bound follows by averaging first over uniform~$U$ and then over the candidate arms.

Subtract the displayed formula for~$Z$ from the actual loss change. Use~$|\ell_{t,C}-\ell_{t,a_s}|\le1$, drop~$\ind_{\{V\le\ell_{s,a_s}\}}$, and average over~$\tau,R$, which are independent of~$\mathcal E_{s,t}$. This gives
\begin{align*}
 \left|\E_b\sum_t(\ell_{t,A_t}-\ell_{t,a_t})-\E_b Z\right|
 &\le\frac1B\sum_{s<t}\Pp(R\ge t-s)\E_b\ind_{\mathcal E_{s,t}^c}\\
 &\le\frac{4m_*}{B^2}\sum_{r=1}^{B-1}(B-r)\Pp(R\ge r)
 \le\frac{4m_*\mu}{B}\,.
\end{align*}

\emph{Cyclic weights.}
For each~$s$,
\begin{equation*}
 \operatorname{Var}_{i\sim p}(\ell_{s,i})
 \le\sum_i p_i(\ell_{s,i}-1/2)^2\le\frac14\,.
\end{equation*}
Cauchy--Schwarz therefore gives
\begin{equation*}
 |\Cov_{i\sim p}(\ell_{s,i},\ell_{t,i})|
 \le\sqrt{\operatorname{Var}_{i\sim p}(\ell_{s,i})
          \operatorname{Var}_{i\sim p}(\ell_{t,i})}\le\frac14\,.
\end{equation*}
Since~$H_{t-s}=H_{t-s}^++H_{B-(t-s)}^+$,
\begin{align*}
 -\frac1{4B}\sum_{s<t}H_{t-s}^+
                      \Cov_{i\sim p}(\ell_{s,i},\ell_{t,i})
 &=-\frac1{4B}\sum_{s<t}H_{t-s}
                      \Cov_{i\sim p}(\ell_{s,i},\ell_{t,i})\\
 &\quad+\frac1{4B}\sum_{s<t}H_{B-(t-s)}^+
                      \Cov_{i\sim p}(\ell_{s,i},\ell_{t,i})\\
 &\le-\frac1{4B}\sum_{s<t}H_{t-s}
                      \Cov_{i\sim p}(\ell_{s,i},\ell_{t,i})
       +\frac1{16B}\sum_{s<t}H_{B-(t-s)}^+\\
 &=-\frac1{4B}\sum_{s<t}H_{t-s}
                      \Cov_{i\sim p}(\ell_{s,i},\ell_{t,i})
       +\frac1{16B}\sum_{r=1}^{B-1}rH_r^+\,.
\end{align*}
Expanding~$H_r^+$ and using~$m_*=2^{J-1}$ now yields
\begin{align*}
 \frac1{16B}\sum_{s<t}H_{B-(t-s)}^+
 &=\frac1{16B}\sum_{j<J}\frac{j+1}{m_j}
                         \sum_{r=1}^{m_j-1}r(1-r/m_j)\\
 &=\frac1{96B}\sum_{j<J}\frac{(j+1)(m_j^2-1)}{m_j}\\
 &\le\frac{J}{96B}\sum_{j<J}m_j
 =\frac{J(2^J-1)}{96B}
 \le\frac{Jm_*}{48B}\,.
\end{align*}
Finally~$m_*=2^{J-1}$ and~$B\ge4^J$ imply
\begin{equation*}
 \frac{\mu}{B}\le\frac1{32},\qquad
 \frac{4m_*\mu}{B}\le\frac3{16},\qquad
 \frac{Jm_*}{48B}\le\frac1{192}\,.
\end{equation*}
These bounds follow from~\eqref{alg:tail}. Adding to the base loss gives
\begin{align*}
 \E_b\sum_t\ell_{t,A_t}-\langle p,D\rangle
 &\le \frac{5}{4}-\frac1{4B}\sum_{s<t}H_{t-s}\Cov_{i\sim p}(\ell_{s,i},\ell_{t,i})\,.
\end{align*}
\end{proof}

\subsection{Bounds for the Tsallis updates}\label{alg:sec-tsallis}

We use the Tsallis potential value and its maximizing distribution
\begin{equation*}
 \Psi(y)=\max_{x\in\Delta_K}\left\{\langle y,x\rangle+2\sum_i\sqrt{x_i}\right\},
 \qquad
 p(y)=\arg\max_{x\in\Delta_K}\left\{\langle y,x\rangle+2\sum_i\sqrt{x_i}\right\}\,.
\end{equation*}
We first bound a sequence of signed updates, then verify the hypotheses and bound the quadratic term for Algorithm~\ref{alg:many}.

We use the stability estimate in Zimmert and Seldin~\cite{tsallis}. We provide a proof here for completeness.

\begin{lemma}[Signed Tsallis updates]\label{sc:tsallis}
Let~$g_1,\ldots,g_Q\in\mathbb R^K$,~$y_1=0$, and~$y_{b+1}=y_b-\eta g_b$ for~$1\le b\le Q$, with~$\eta>0$ and~$p_b=p(y_b)$. Suppose that each update satisfies
\begin{equation}\label{sc:sparse-step}
\eta\max_{a\in[K]}|g_{b,a}|\sqrt{p_{b,a}}\le\frac1{3}\,.
\end{equation}
Then for every coordinate vector~$e_i$,
\begin{equation}\label{sc:tsallis-regret}
 \sum_{b=1}^Q\langle p_b-e_i,g_b\rangle
 \le\frac{2\sqrt K}{\eta}
       +\eta\sum_{b=1}^Q\sum_{a=1}^K
        \frac{p_{b,a}^{3/2}g_{b,a}^2}
             {1+\eta g_{b,a}\sqrt{p_{b,a}}}\,.
\end{equation}
\end{lemma}

\begin{proof}
The maximum defining~$p(y)$ exists by continuity and compactness and is unique by strict concavity. It lies in the interior of $\Delta_K$, since moving mass~$\epsilon$ to a zero coordinate increases the square-root term by~$2\sqrt\epsilon$, while changing the other terms by~$O(\epsilon)$. For a fixed block, write~$y,g,p$ for~$y_b,g_b,p_b$. The Lagrangian multiplier condition gives $y_i+p_i^{-1/2}=\lambda$ for some $\lambda$, therefore
\begin{equation*}
 \Psi(y)=\sum_i(y_ip_i+2\sqrt{p_i})
 =\lambda+\sum_i\sqrt{p_i}\,.
\end{equation*}
By~\eqref{sc:sparse-step}, we have~$p_i^{-1/2}+\eta g_i=p_i^{-1/2}(1+\eta g_i\sqrt{p_i})>0$. Therefore, for any~$z\in\Delta_K$,
\begin{align*}
 \langle y-\eta g,z\rangle+2\sum_i\sqrt{z_i}
 &=\lambda+\sum_i\left[-(p_i^{-1/2}+\eta g_i)z_i+2\sqrt{z_i}\right]\\
 &\le\lambda+\sum_i\frac{\sqrt{p_i}}{1+\eta g_i\sqrt{p_i}}\,.
\end{align*}
Maximizing over~$z$ and subtracting~$\Psi(y)$ gives
\begin{align*}
 \Psi(y-\eta g)-\Psi(y)+\eta\langle p,g\rangle
 &\le\sum_i\left[
      \frac{\sqrt{p_i}}{1+\eta g_i\sqrt{p_i}}
       -\sqrt{p_i}+\eta p_i g_i\right]\\
 &=\eta^2\sum_i\frac{p_i^{3/2}g_i^2}{1+\eta g_i\sqrt{p_i}}\,.
\end{align*}
Summing over blocks yields
\begin{equation*}
 \Psi(y_{Q+1})-\Psi(0)
 \le-\eta\sum_{b=1}^Q\langle p_b,g_b\rangle
       +\eta^2\sum_{b=1}^Q\sum_a
       \frac{p_{b,a}^{3/2}g_{b,a}^2}
            {1+\eta g_{b,a}\sqrt{p_{b,a}}}\,.
\end{equation*}
Cauchy--Schwarz gives~$\Psi(0)=2\sqrt K$. Evaluating the maximum at~$e_i$ gives~$\Psi(y_{Q+1})\ge2+\langle y_{Q+1},e_i\rangle=2-\eta\sum_{b=1}^Q g_{b,i}$. Substitution and division by~$\eta$ prove~\eqref{sc:tsallis-regret}.
\end{proof}

\paragraph{Normalization and block updates.}
\begin{lemma}[Comparison normalization]\label{alg:normalization}
If~$\alpha\in(0,1)$,~$k\in\{1,\ldots,B-1\}$, and~$|k/B-\alpha|\le1/B$, then
\begin{equation}\label{alg:balance}
 S_k^2\ge\frac5{16}\sqrt{\min\{\alpha,1-\alpha\}}\log^2 B\,.
\end{equation}
\end{lemma}

\begin{proof}
Put~$k'=\min\{k,B-k\}$ and~$\beta=\min\{\alpha,1-\alpha\}$. If~$k'\ge m_*$, then~\eqref{alg:normalizing-mass} gives~$S_k=J$. Since~$M\le2J+1\le3J$ and~$\beta\le1/2$,
\begin{equation*}
 \frac{\sqrt\beta\log^2 B}{S_k^2}
 \le\frac{9(\log2)^2}{\sqrt2}<\frac{16}{5}\,.
\end{equation*}
If~$k'<m_*$, put~$u=1+\lfloor\log_2 k'\rfloor$. The first~$u$ scales in~\eqref{alg:normalizing-mass} each contribute one, so~$S_k\ge u$. The rounding hypothesis and integrality of~$k'$ give
\begin{equation*}
 \beta\le\frac{k'+1}{B}\le2^{u-M}\,.
\end{equation*}
Since~$1\le u\le J-1<M$, we have~$M/u\le M-u+1$. Hence
\begin{align*}
 \frac{\beta^{1/4}\log B}{S_k}
 &\le\frac{M\log2}{u}\,2^{(u-M)/4}\\
 &\le(M-u+1)(\log2)2^{-(M-u)/4}\\
 &\le\max_{x\ge0}(x+1)(\log2)2^{-x/4}
 =\frac{4\,2^{1/4}}{\mathrm e}<\frac4{\sqrt5}\,,
\end{align*}
since the maximum occurs at~$x=4/\log2-1$. Squaring proves~\eqref{alg:balance}.
\end{proof}

\begin{lemma}[Block update estimates]\label{alg:pair-update}
In a fixed block of Algorithm~\ref{alg:many}, the update~$g$ is bounded and satisfies~\eqref{sc:sparse-step}. Then for every arm~$i$,
\begin{equation}\label{alg:many-unbiased}
 \E_b\langle p-e_i,g\rangle=\langle p,D\rangle-D_i\,.
\end{equation}
The quadratic term satisfies
\begin{equation}\label{alg:positive-covariance}
 \eta\E_b\sum_{a=1}^K
       \frac{p_a^{3/2}g_a^2}{1+\eta g_a\sqrt{p_a}}
 \le\frac{27}{100}+\frac{1}{4B}\sum_{s<t}H_{t-s}
                                  \Cov_{i\sim p}(\ell_{s,i},\ell_{t,i})\,.
\end{equation}
\end{lemma}

\begin{proof}
\emph{Mean.} Set~$\omega_{ji}=\omega_{ij}$ and~$\widehat D_{ji}=-\widehat D_{ij}$ when reversing the pair. On a distinct pair, the two nonzero coordinates can only be~$g_i=p_j\widehat D_{ij}/\omega_{ij}$ and~$g_j=-p_i\widehat D_{ij}/\omega_{ij}$. These give~$\langle p,g\rangle=0$, as does the zero update on a singleton. Averaging over pairs using~\eqref{alg:mean-range} gives
\begin{equation*}
 \E_b g_i
 =\sum_{j\ne i}\omega_{ij}\frac{p_j}{\omega_{ij}}(D_i-D_j)
 =D_i-\langle p,D\rangle\,.
\end{equation*}
Thus~$\E_b\langle p-e_i,g\rangle=-\E_b g_i$, which gives~\eqref{alg:many-unbiased}.

\emph{Step size.} On a distinct pair, use~$|\widehat D_{ij}|\le BJ$ from~\eqref{alg:mean-range} and~$\omega_{ij}=p_ip_j+(p_i+p_j)/(2K)\ge p_j(p_i+1/(2K))$. The arithmetic-geometric mean inequality gives
\begin{equation*}
 |g_i|\sqrt{p_i}
 \le BJ\frac{\sqrt{p_i}}{p_i+1/(2K)}
 \le BJ\sqrt{K/2}\,.
\end{equation*}
The same bound holds for~$j$, and singleton updates are zero. Also~$\omega_{ij}\ge p_j/(2K)$ gives~$|g_i|\le2KBJ$, and likewise for~$j$. With~$\eta=\log^2B/(432\sqrt{K}B^2)$ and~$J\le \log B/(2\log 2)$, we obtain
\begin{equation*}
 \eta\max_a|g_a|\sqrt{p_a}
 \le \frac{\log^3 B}{864\sqrt 2(\log 2)B}
 \le\frac{27}{864\sqrt 2(\log 2)\mathrm e^3}<\frac1{81}\,,
\end{equation*}
where we use the fact that~$x^3e^{-x}$ is maximized at~$x=3$.

\emph{Pair coefficient.} Fix a distinct pair~$i<j$. Put~$a=\max\{p_i,p_j\}$,~$c=\min\{p_i,p_j\}$, and~$\beta=\min\{\alpha_{ij},1-\alpha_{ij}\}$. Then~$\omega_{ij}=ac+(a+c)/(2K)$ and
\begin{equation*}
 \beta=\frac{c(a+1/K)}{2ac+(a+c)/K}\,.
\end{equation*}
The identity
\begin{align*}
 &(a+1/K)(2ac+(a+c)/K)-\frac aK(a+c+2\sqrt{ac})\\
 =& 2a\left(\sqrt{ac}-\frac1{2K}\right)^2
          +\frac{a/2+c}{K^2}+\frac{2ac}{K}\ge0
\end{align*}
implies
\begin{equation*}
 \left(\frac{\sqrt{p_i}p_j+\sqrt{p_j}p_i}{\omega_{ij}}\right)^2
 =\frac{ac(a+c+2\sqrt{ac})}{\omega_{ij}^2}
 \le 4K\beta\,.
\end{equation*}
Rounding gives~$|k_{ij}/B-\alpha_{ij}|\le1/B$, so Lemma~\ref{alg:normalization} applies. Take the square root and use~\eqref{alg:balance} to obtain
\begin{equation}\label{alg:pair-balance}
 \frac{\sqrt{p_i}p_j+\sqrt{p_j}p_i}{\omega_{ij}S_{k_{ij}}^2}
 \le\frac{32\sqrt K}{5\log^2 B}\qquad(i<j)\,.
\end{equation}

\emph{Quadratic term.}
On~$\{i,j\}$ with~$i<j$, the update has only the coordinates~$p_j\widehat D_{ij}/\omega_{ij}$ and~$-p_i\widehat D_{ij}/\omega_{ij}$. The pair occurs with probability~$\omega_{ij}$, and the update on a singleton is zero. Hence
\begin{equation}\label{alg:many-charge}
 \E_b\sum_{a=1}^K p_a^{3/2}g_a^2
 =\sum_{i<j}p_ip_j
    \frac{\sqrt{p_i}p_j+\sqrt{p_j}p_i}{\omega_{ij}}\,
           \E_b[\widehat D_{ij}^{\,2}\mid\{i,j\}]\,.
\end{equation}
The step-size estimate gives~$1+\eta g_a\sqrt{p_a}\ge80/81$. Apply~\eqref{alg:pair-balance} to the nonnegative second moments, then insert the signed estimate~\eqref{alg:moment}. This gives
\begin{align*}
 \eta\E_b\sum_{a=1}^K
      \frac{p_a^{3/2}g_a^2}{1+\eta g_a\sqrt{p_a}}
 &\le\frac{81\eta}{80}\E_b\sum_{a=1}^Kp_a^{3/2}g_a^2\\
 &\le\frac{162\eta\sqrt K}{25\log^2 B}
   \sum_{i<j}p_ip_j S_{k_{ij}}^2
      \E_b[\widehat D_{ij}^{\,2}\mid\{i,j\}]\\
 &\le\frac{27}{50}\sum_{i<j}p_ip_j
    +\frac1{4B}\sum_{s<t}H_{t-s}\sum_{i<j}
      p_ip_j(\ell_{s,i}-\ell_{s,j})(\ell_{t,i}-\ell_{t,j})\,.
\end{align*}
In the last line the common coefficient is~$162\eta\sqrt K/(25\log^2 B)=3/(200B^2)$. Multiplication by the two coefficients~$36B^2$ and~$50B/3$ from~\eqref{alg:moment} gives~$27/50$ and~$1/(4B)$, respectively. Also~$\sum_{i<j}p_ip_j=(1-\sum_i p_i^2)/2\le1/2$. For the remaining sum, include both orders of each pair and expand:
\begin{align*}
 \sum_{i<j}p_ip_j(\ell_{s,i}-\ell_{s,j})(\ell_{t,i}-\ell_{t,j})
 &=\tfrac12\sum_{i,j}p_ip_j(\ell_{s,i}-\ell_{s,j})(\ell_{t,i}-\ell_{t,j})\\
 &=\sum_i p_i\ell_{s,i}\ell_{t,i}
             -\langle p,\ell_s\rangle\langle p,\ell_t\rangle\\
 &=\Cov_{i\sim p}(\ell_{s,i},\ell_{t,i})\,.
\end{align*}
This proves~\eqref{alg:positive-covariance}.
\end{proof}

\subsection{Proof of the regret bound}\label{alg:sec-regret}

\begin{proof}[Proof of the upper bound in Theorem~\ref{thm:cost}]
Fix a dyadic block length~$4\le B\le T$ and run Algorithm~\ref{alg:many} on the zero-padded table with~$Q=\lceil T/B\rceil$ blocks. Choose~$i_\star$ to minimize~$\sum_{b=1}^Q D_{b,i}$. By Lemma~\ref{alg:pair-update}, the updates are bounded and obey the step-size condition in Lemma~\ref{sc:tsallis}. Take expectations in~\eqref{sc:tsallis-regret} with~$i=i_\star$, use~\eqref{alg:many-unbiased}, and add the difference between actual and comparison loss to obtain
\begin{align}\label{alg:regret-decomposition}
 R_T(A,\ell)
 \le\frac{2\sqrt K}{\eta}
 +\sum_{b=1}^Q\E\left[
     \E_b\sum_{t=0}^{B-1}\ell_{b,t,A_{b,t}}
     -\langle p_b,D_b\rangle
     +\eta\E_b\sum_{a=1}^K
       \frac{p_{b,a}^{3/2}g_{b,a}^2}
            {1+\eta g_{b,a}\sqrt{p_{b,a}}}
                         \right]\,.
\end{align}
Padding leaves regret unchanged because every arm has zero loss on the added rounds. In each block, add~\eqref{alg:clock-loss} and~\eqref{alg:positive-covariance}. Their covariance sums cancel, so the bracket in~\eqref{alg:regret-decomposition} is at most~$5/4+27/100=38/25$.

The base path has at most $2$ switches within a block. At most one detour is triggered, and it replaces a single interval on which the base arm is constant. Its entrance and exit can add at most $2$ switches, giving at most $4$ switches within each block. There is at most one switch at each of the~$Q-1$ block boundaries. Truncating the padded path can only remove switches. Therefore~$S_T(A,\ell)\le4Q+(Q-1)\le5Q$, and
\begin{equation}\label{alg:many-bound}
 R_T(A,\ell)+S_T(A,\ell)
 \le 864\frac{KB^2}{\log^2 B}+\frac{163}{25}\left\lceil\frac{T}{B}\right\rceil\,.
\end{equation}

Set~$x=T/K$ and~$u=(x\log^2(2+x))^{1/3}$. If~$x\ge\mathrm e^{12}$, take $B=2^{\lceil\log_2(u/24)\rceil}$. Then
\begin{equation*}
 \log B\ge\frac{1}{3}\log x + \frac{2}{3}\log 12 - \log 24\ge\frac{1}{5}\log (2+x)\,.
\end{equation*}
The two terms in~\eqref{alg:many-bound} therefore satisfy
\begin{align*}
 864K\frac{B^2}{\log^2 B}
 &\le\frac{864\cdot25}{12^2}\frac{T}{u}
 =150\frac{T}{u},\\
 \frac{163}{25}\left\lceil\frac{T}{B}\right\rceil
 &\le\frac{163}{25}\left(\frac{24T}{u}+1\right)
 \le160\frac{T}{u}\,.
\end{align*}
Hence~$R_T(A,\ell)+S_T(A,\ell)\le310T/u$. 

If~$x<\mathrm e^{12}$, play a fixed arm. Its cost is at most~$T$, and~$u\le\mathrm e^4 13^{2/3}<310$, so the same bound holds. Choosing the better guarantee from this rule and a fixed arm gives
\begin{equation}\label{alg:uniform-bound}
 R_T(A,\ell)+S_T(A,\ell)
 \le\min\left\{T,\,
 \frac{310K^{1/3}T^{2/3}}{\log^{2/3}(2+T/K)}\right\}\,.
\end{equation}
\end{proof}

%% file: bandit-lower.tex
\section{Refining the multiscale lower bound}\label{sec:bandit-lower}

By Yao's minimax principle, it suffices to construct a distribution depending only on~$T,K$ under which every deterministic learner has the required expected cost. For~$K\le T$, we restart the multiscale walk of~\cite{ddkp} in~$K$ blocks and clip the resulting loss sequences into $[0,1]$.

\subsection{The loss experiments}
Suppose~$2\le K\le T$. Partition the horizon into~$K$ nonempty consecutive blocks of length at most~$H=\lceil T/K\rceil$. Set
\begin{equation}\label{lower:depth}
 d=1+\lfloor\log_2H\rfloor,\qquad \sigma^2=\frac1{64d}\,.
\end{equation}
In block~$j$, generate the Gaussian walk independently of the other blocks using the recursion
\begin{equation}\label{lower:forest}
 W_{j,0}=0,\qquad W_{j,r}=W_{j,\rho(r)}+\xi_{j,r},
 \qquad \rho(r)=r-2^{v_2(r)}\,,
\end{equation}
where~$r$ is the local round starting at~$1$,~$v_2(r)=\max\{q\in\mathbb Z_{\ge0}:2^q\mid r\}$, and the increments~$\xi_{j,r}$ are independent~$N(0,\sigma^2)$. Let~$\clip(z)=\min\{1,\max\{0,z\}\}$. Fix~$0<\varepsilon\le1/16$. The null experiment~$Q_0$ gives all arms loss~$\clip(1/2+W_{j,r})$. In the hidden-arm experiment~$Q_i$, for~$i\in[K]$,
\begin{equation}\label{lower:losses}
 \ell_{j,r,a}=\clip\bigl(1/2+W_{j,r}-\varepsilon\ind_{\{a=i\}}\bigr)\,.
\end{equation}

\begin{lemma}\label{lower:information}
Fix a deterministic learner $\mathcal{A}$. Let~$m=\E_{\ell\sim Q_0}S_T(\mathcal A,\ell)$. Then
\begin{equation}\label{lower:regret}
 \frac1K\sum_{i=1}^K\E_{\ell\sim Q_i}R_T(\mathcal A,\ell)
 \ge\varepsilon T\left(\frac{5}{12}-4\varepsilon d\sqrt{\frac{2m+K}{K}}\right)\,.
\end{equation}
\end{lemma}

\begin{proof}
Write~$\E_i$ for expectation under~$Q_i$, including~$i=0$. Let~$M$ be the realized number of switches, so~$m=\E_0M$.

\emph{Divergence.} By~\cite[Lemma~2]{ddkp}, each block has depth and width at most~$d$. In particular,~$\operatorname{Var}(W_{j,r})\le d\sigma^2=1/64$ for every $j$ and $r$. There are at most~$d$ edges from the root in each block, since~$\rho(r)=0$ precisely for powers of two.

Let~$Q_i^{\rm raw}$ and~$Q_i^{\rm obs}$ be the laws of the selected raw losses and clipped observations, respectively. Put~$I^i_{j,r}=\ind_{\{A_{j,r}=i\}}$ and~$I^i_{j,0}=0$. The divergence chain-rule calculation in the proof of~\cite[Lemma~3]{ddkp}, applied conditionally on the observations preceding each block, gives
\begin{equation}\label{lower:kl}
 \KL(Q_0^{\rm raw}\Vert Q_i^{\rm raw})
 =\frac{\varepsilon^2}{2\sigma^2}
       \E_0\sum_{j,r}(I^i_{j,r}-I^i_{j,\rho(r)})^2\,.
\end{equation}
The learner's actions are deterministic functions of its clipped observations, which are functions of its raw observations. Thus the same divergence calculation applies although the learner observes only clipped losses.

If~$\rho(r)=0$, then~$I^i_{j,0}=0$ for every~$i$, so
\begin{equation*}
 \sum_{i=1}^K(I^i_{j,r}-I^i_{j,0})^2
 =\sum_{i=1}^K\ind_{\{A_{j,r}=i\}}=1\,.
\end{equation*}
If~$\rho(r)>0$, then
\begin{align*}
 \sum_{i=1}^K(I^i_{j,r}-I^i_{j,\rho(r)})^2
 &=\sum_{i=1}^K\bigl(I^i_{j,r}+I^i_{j,\rho(r)}-2I^i_{j,r}I^i_{j,\rho(r)}\bigr)=2-2\ind_{\{A_{j,r}=A_{j,\rho(r)}\}}\\
 &=2\ind_{\{A_{j,r}\ne A_{j,\rho(r)}\}}
 \le2\sum_{s=\rho(r)+1}^{r}\ind_{\{A_{j,s}\ne A_{j,s-1}\}}\,.
\end{align*}
For fixed~$j$ and~$q\in\{0,\ldots,d-1\}$, the intervals~$(\rho(r),r]$ with~$v_2(r)=q$ have length~$2^q$ and distinct left endpoints that are multiples of~$2^{q+1}$. They are therefore disjoint. Each switch is consequently counted at most once per scale, hence at most~$d$ times. Since there are at most~$Kd$ root edges,
\begin{align*}
 \sum_{i=1}^K\sum_{j,r}(I^i_{j,r}-I^i_{j,\rho(r)})^2
 &\le Kd+2\sum_{j,r:\rho(r)>0}\sum_{s=\rho(r)+1}^{r}
       \ind_{\{A_{j,s}\ne A_{j,s-1}\}}\\
 &\le Kd+2dM\,.
\end{align*}
Data processing, Pinsker's inequality and Cauchy--Schwarz yield
\begin{align}\label{lower:tv}
 v&:=\frac1K\sum_i\TV(Q_i^{\rm obs},Q_0^{\rm obs})
 \le \sqrt{\frac1{2K}\sum_i\KL(Q_0^{\rm raw}\Vert Q_i^{\rm raw})}\notag\\
 &\le\varepsilon\sqrt{\frac{d(2m+K)}{4K\sigma^2}}
 =4\varepsilon d\sqrt{\frac{2m+K}{K}}\,,
\end{align}
where the sums over hidden arms run from~$1$ to~$K$. Let~$N_i$ count plays of arm~$i$. The occupation argument in the proof of~\cite[Lemma~5]{ddkp} uses~$0\le N_i\le T$ and~$\sum_i\E_0N_i=T$ to give
\begin{equation}\label{lower:occupancy}
 \frac1K\sum_i\E_iN_i\le\frac{T}{K}+Tv\,.
\end{equation}

\emph{Clipping.} 
Fix~$i\in[K]$ and define the clipped loss gap at round~$(j,r)$ by
\begin{equation*}
 G_{j,r}=\clip(1/2+W_{j,r})
              -\clip(1/2+W_{j,r}-\varepsilon)\,.
\end{equation*}
The clipping map is increasing and~$1$-Lipschitz, so~$0\le G_{j,r}\le\varepsilon$. If~$|W_{j,r}|\le7/16$, both arguments lie in~$[0,1]$ because~$\varepsilon\le1/16$, and~$G_{j,r}=\varepsilon$. Hence, pointwise,
\begin{equation*}
 0\le\varepsilon-G_{j,r}
 \le\varepsilon\ind_{\{|W_{j,r}|>7/16\}}\,.
\end{equation*}
Since~$\E_iW_{j,r}=0$ and~$\operatorname{Var}(W_{j,r})\le1/64$, Chebyshev's inequality gives, separately at every round,
\begin{equation*}
 \E_i(\varepsilon-G_{j,r})
 \le\varepsilon\Pp(|W_{j,r}|>7/16)
 \le\frac{\varepsilon/64}{(7/16)^2}
 =\frac{4\varepsilon}{49}\,.
\end{equation*}
To use this unconditional estimate despite the dependence between~$A_{j,r}$ and~$W_{j,r}$, write
\begin{align*}
 G_{j,r}(1-I^i_{j,r})
 &=\varepsilon(1-I^i_{j,r})
       -(\varepsilon-G_{j,r})(1-I^i_{j,r})\\
 &\ge\varepsilon(1-I^i_{j,r})-(\varepsilon-G_{j,r})\,.
\end{align*}
Arm~$i$ is best at every round under~$Q_i$. Taking expectations and summing therefore yields
\begin{align*}
 \E_{\ell\sim Q_i}R_T(A,\ell)
 &=\E_i\sum_{j,r}G_{j,r}(1-I^i_{j,r})
 \ge\varepsilon(T-\E_iN_i)-\frac{4\varepsilon T}{49}\,.
\end{align*}
Averaging over~$i$ and applying~\eqref{lower:occupancy},
\begin{equation*}
 \frac1K\sum_{i=1}^K\E_{\ell\sim Q_i}R_T(A,\ell)
 \ge\varepsilon T\left(\frac{45}{49}-\frac1K-v\right)
 \ge\varepsilon T\left(\frac{5}{12}-v\right)\,.
\end{equation*}
Substituting~\eqref{lower:tv} proves~\eqref{lower:regret}.
\end{proof}

\subsection{From information to switching cost}
\begin{proof}[Lower bound in Theorem~\ref{thm:cost}]
For~$K\le T$, set
\begin{equation*}
 \Gamma=\left(\frac{KT^2}{d^2}\right)^{1/3},
 \qquad \varepsilon=\frac{\Gamma}{24T}\le\frac1{16}\,.
\end{equation*}
Write~$m=\E_{\ell\sim Q_0}S_T(A,\ell)$. Since~$d\le H\le2T/K$ and~$2^{2/3}<8/5$, we have
\begin{align*}
 4\varepsilon d\sqrt{\frac{2m+K}{K}}
 &=\frac16\sqrt{\left(\frac{Kd}{T}\right)^{2/3}+\frac{2m}{\Gamma}}\\
 &\le\frac1{12}\left(1+\left(\frac{Kd}{T}\right)^{2/3}+\frac{2m}{\Gamma}\right)\\
 &\le\frac{13}{60}+\frac{m}{6\Gamma}\,.
\end{align*}
Lemma~\ref{lower:information} therefore yields
\begin{align*}
 \frac1K\sum_{i=1}^K\E_{\ell\sim Q_i}R_T(A,\ell)
 &\ge\frac{\Gamma}{24}\left(
       \frac{11}{12}-\frac12-\frac{13}{60}-\frac{m}{6\Gamma}\right)\\
 &=\frac{\Gamma}{120}-\frac{m}{144}\,.
\end{align*}
Under~$Q_0$, expected cost equals~$m$; under every~$Q_i$, switching costs are nonnegative. Therefore,
\begin{align*}\label{lower:prior}
 \sup_\ell\bigl(R_T(A,\ell)+S_T(A,\ell)\bigr)
 &\ge\frac{m}{145}
       +\frac{144}{145K}\sum_{i=1}^K\E_{\ell\sim Q_i}R_T(A,\ell)\\
 &\ge\frac{m}{145}
       +\frac{144}{145}\left(\frac{\Gamma}{120}-\frac{m}{144}\right)\\
 &=\frac{6\Gamma}{725}\ge
 \frac{K^{1/3}T^{2/3}}{200\log^{2/3}(2+T/K)}\,.
\end{align*}

For~$K\ge T$, the standard argument gives a linear regret lower bound.
\end{proof}